# A Neural Model of Rule Discovery with Relatively Short-Term Sequence Memory

Naoya Arakawa[1]

## Abstract

This report proposes a neural cognitive model for discovering regularities in event sequences. In a fluid intelligence task, the subject is required to discover regularities from relatively short-term memory of the first-seen task. Some fluid intelligence tasks require discovering regularities in event sequences. Thus, a neural network model was constructed to explain fluid intelligence or regularity discovery in event sequences with relatively short-term memory. The model was implemented and tested with delayed match-to-sample tasks.

## 1. Introduction

The objective of this report is to propose a neural-network-based cognitive model for discovering regularities in event sequences. In a fluid intelligence task such as a visual analogy task, subjects are required to identify regularities in the task that is novel to them from their relatively brief memories of the encountered situations. In tasks that involve the presentation of event sequences, subjects are required to use the memories to discover the regularity in the sequences. Even in tasks such as visual analogy tasks that do not explicitly involve event sequences, the agent has to deal with event sequences as they should involve sequential eye movement and attentional selection.

For a cognitive model to be biologically plausible, it must be realizable by a neural network. In the problem setting above, "memory" for relatively short periods of time within the task is used rather than "learning" by long-term modification of neural circuits. As the neural mechanisms responsible for memory for relatively short periods of time, short-term enhancement of synaptic plasticity (weight) and/or recurrent activity can be thought of. Below, a neural circuit model using short-term weight enhancement is proposed, and its implementation and evaluation are reported. The objective of this report is to show that a relatively simple neural network model can solve a test task of fluid intelligence, and the proposed model may not correspond to actual physiological or psychological measurements, nor aim at precise biological plausibility.

This report is based on the following assumptions: 1) the agent has an input-output sequence and a corresponding internal state sequence; 2) the agent can recall past internal state and input-output sequences from the current input sequence; 3) the agent can replay the internal state sequence in reverse when a reward is given, in order to evaluate the sequence. 4) the agent probabilistically determines the output using parallelly recalled internal state sequences and the values associated to the sequences; 5) the agent pays attention to a specific attribute in the input at a time and remembers only the selected attribute. The rationale behind these assumptions will be explained in the chapter of Method.

---

[1] The Whole Brain Architecture Initiative, naoya.arakawa@nifty.com

The model was tested with the delayed match-to-sample task, which is typically used to test short-term or working memory, and requires the discovery of regularities in a given sequence of events in order to understand what the task is about.

## 2. Background
### 2.1. Fluid Intelligence

Psychometrics, a branch of psychology, has been concerned with the measurement of human intelligence and with what factors are involved in intelligence. Even in studies based on the hypothesis that human intelligence can be attributed to a single general intelligence factor *g* (Spearman, 1904), multiple factors have been postulated. Cattell (1943) posited a difference between fluid and crystallized intelligence in general intelligence. According to (Cattell, 1963), crystallized intelligence is of 'skilled judgment habits,' and fluid intelligence is of 'adaptation to new situations, where crystallized skills are of no particular advantage.'

Fluid intelligence is a 'factor' obtained by the factor analysis of psychometric tests, and certain types of tests are statistically related to this factor. Specifically, visual analogy/reasoning tasks such as Raven's Progressive Matrices are considered to be closely related to fluid intelligence (Carpenter, 1990). The visual analogy task requires the discovery of common relationships between multiple figures. What is required here can be described as the *discovery of regularity*. (Incidentally, Chuderski (2022) stated that fluid intelligence did not require the discovery of rules, however it seems rather to be concerned with the discovery of rules across multiple tests.)

In general, fluid intelligence requires working memory (short-term memory for solving a task) (Hagemann, 2023)(Unsworth, 2014). Even when a task is given at once, as in a visual analogy task, the agent has to pay attention to one of the elements in the task at a time, compare a presented element with another memorized element, to find relationships among them. Thus, memory is needed to handle multiple elements.

One of the working memory tasks is the delayed match-to-sample task, where a sample figure is presented, and after the sample figure is hidden, target figures are presented. Subjects select a figure that is considered to be the same as the sample according to certain criteria. The subject must remember the attributes of the sample figure when selecting the target figure. Delayed match-to-sample tasks also require the discovery of rules (unless the rules are given linguistically beforehand). Subjects must discover the rule of selecting a figure that is considered the same as the sample, as well as the rule regarding the criterion to be used for comparison. In this respect, the delayed match-to-sample task can also be regarded as a fluid intelligence task. The regularity in the delayed match-to-sample task is to be found in the input sequence.

## 2.2. Prior Studies

Various models have been proposed for solving the visual analogy task (Małkiński, 2022), which is considered highly relevant to fluid intelligence and is also interpreted as a rule discovery process (Chollet, 2019). Early studies used rule-based methods, followed by analogy discovery methods, and more recently, machine learning has been used. Logical methods such as inductive logical programming have also been used for research in rule discovery (Cropper, 2022). Although these studies are interesting, they diverge from the interest of this report, which is the discovery of sequence rules by neural circuits. Though recent methods with machine learning use artificial neural circuits, they are based on statistical learning that requires a large number of trials and thus cannot be regarded as a model that can discover the rules of a task on the first try (zero shot).

Working memory is considered necessary for fluid intelligence. As a neural model of working memory, the prefrontal cortex basal ganglia working memory (PBWM) model was proposed (O'Reilly, 2006). While It is a mechanism for learning what to remember in response to a task, it is not a model for discovering rules. However, as there is an implementation of sequence memory using PBWM (O'Reilly, 2002), it may show similar performance to the model in this proposal. There is also a working memory model with RNNs (Xie, 2022). While Amit (2013) attempted to handle the delayed match-to-sample task with a relatively simple neural network model, it is rather a model of shape recognition.

## 3. Method
## 3.1. Sequence Memory

As noted above, fluid intelligence tasks may require sequence memory. This report thus addresses tasks that require sequence representations for solution. Sequence memory here remembers/memorizes a sequence with a single exposure; it is not a statistical learner that learns sequences with multiple exposures. Sequencel memory cannot be attained simply by associating attributes at one time with the next ones. This is because an input sequence may contain transitions from the same attributes to several different attributes; i.e., transitions may be one to many in the attribute space. This problem can be solved by using latent state sequences having one-to-one transition, where each of the states is associated with attributes to be remembered. This leads to Assumption 1) having a sequence of internal states corresponding to the input-output sequence. When these latent states are represented in a linear space (vector), each state must be independent in order to be correctly recalled (or sequence recall may be mixed up). Memory in vector spaces or neural circuits has been studied with the *associatron* (Nakano, 1972), followed by the Hopfield model (Hopfield, 1982), Morita's more biologically plausible model (Morita, 1993), and more recently Modern Hopfield Networks (Dense Associative Memories) (Krotov, 2016), which succeeded in greatly increasing the number of patterns a circuit can hold. Sequence memories using such associative memories have also been proposed (Morita 1996)(Chaudhry, 2023).

The present report requires a biologically plausible and simple model that meets the specification of sequence memory. The following is a description of a sequence memory model used in this report. Note that this model is not the main point of this report, and other models that satisfy the specifications may be used. The competitive queuing model (Bullock, 2003)(Houghton, 1990)(Burgess, 1999) is such a model and similar to the model proposed in this report. Besides, O 'Reilly (2002) proposed a neural circuit sequence memory model that did not use competitive queuing (or synaptic memory) and (Raju, 2024) a graph-structured sequence memory.

The sequence memory in this report used a one-hot-vector as the internal (latent) state of a sequence. The one-hot-vectors represented mutually independent states of dimension N. Each cell (neuron) had an activity value, and the one with the largest activity value "fires" as one-hot (as in the key-value memory in (Tyulmankov, 2021)). Though the cell with the maximum value could be identified by a winner-takes-all mechanism with a mutual inhibitory circuit, simple argmax was used in the implementation. The transition between internal states was specified by an association matrix (N x N dimensions) between the cells representing the internal states. Two other associative matrices were also set up between input attributes and internal states. These associations could be biologically realized by short-term (Hebbian) potentiation in the brain.

The activity of the cells was set to decrease with time. This allowed cells to have temporal information, so that cells that had not been used recently could be "recycled." Though the decrease in activity could be achieved, for example, through reciprocal coupling circuits and short-term potentiation, a simple exponential decay was employed in the implementation. The least active cell (for recycling) could also be found with a reciprocal inhibition circuit (by performing lateral inhibition among cells inhibited by the least active cells). However, again, the implementation used a simple argmin for that purpose.

## 3.2. Rules

A rule or hypothesis had the format:

{attribute}* ⇒ action, attributes (prediction)

That is, the action and input attributes at time t+1 were predicted from the input attribute sequence up to time t.

In the implementation, only action was attention to select an attribute (see below). Though it would be generally better to have action type in the antecedent, as there was only one type of action (i.e., attention) in the implementation, only attributes were used in the antecedent.

## 3.3. Mechanism for Rule Discovery

The mechanism is summarized as below:

- The agent receives a sequence of input attributes and determines action at each time.
- The agent memorizes all the sequence of attended input attributes and actions.
- For a new sequence, the agent recalls zero or more past sequences (positive hypothetical sequences; see below) that match the input attribute sequence, and if one or more past sequences match to the input, it predicts the input attributes and an action for the next time step (Assumption 2). The agent recalls and retains multiple sequences, as it is not possible to determine which past sequence will match at the initial stages. (An alternative approach would use a tree or graph structure for sequence representation, or backtrack, but to complicate the mechanism.)
- The success or failure is determined when the agent predicts the last attended input attribute of the episode. A past sequence that succeeds in predicting the last input is registered as a **positive hypothetical sequence**, and a past sequence that fails to predict the last input is marked as a **negative hypothetical sequence**. The cells representing the internal state of the sequence are given (associated to) "reinforcement values"; each internal state cell of a successful or unsuccessful sequence is assigned a positive or negative value, respectively. For the value assignment, a reverse replay with an association matrix is performed from the last state of the sequence to the first state (Assumption 3). Reverse replay was used to avoid a larger number of trials incurred with value assignment algorithms in regular reinforcement learning.
- The action at each time step is determined by the actions recorded in the positive hypothetical sequences that matched the input up to the step. Specifically, if multiple types of actions are found in the matched positive hypothesis sequence, the action is determined from their numbers with multinomial distribution (or a dice throw) (Assumption 4). Since a positive hypothetical sequence is only a hypothesis (i.e., it may have been successful by chance), the action is chosen in a probabilistic manner (it does not have to be by multinomial distribution, though).

## 3.4. Attention as Action

The human cognitive system processes attended attributes in order to reduce information complexity, and no conscious memory takes place without attention (Assumption 5). This cognitive feature is called *selective attention* (van Moorselaar, 2020). In this respect, attention is considered to be action (cf. (Parr, 2014)). In the current experiment, the only action in the model used was attention. When there were multiple salient targets, the target of attention was either chosen randomly or by the action decision mechanism described above, where the only input attributes stored in the sequence memory were those selected by attention.

## 4. Task

A delayed match-to-sample task was used. Each episode consisted of the four phases in the following order: 1) start cue presentation period, 2) sample presentation period, 3) target presentation period, and 4) answer period. The input (observation – environmental output) was a concatenation of the attributes listed below (eight binary digits). The attribute to be attended for matching was one of the two Input attributes, the other being dummy (spurious).

- Start cue (task switch): a one-or-zero-hot-vector representing the start of an episode and the input attribute to be attended.
  If [1, 0], the input attribute to be attended was the Input attribute 1, and if [0, 1], the Input attribute 2 (see below).
  It was [0, 0] except for the start cue presentation period.
- Input attribute 1: one-or-zero-hot-vector
  It was a one-hot-vector during the sample and target presentation periods and a zero vector during the rest of the time.
  If the start cue was [1, 0], it took the value used for matching. Otherwise, the one-hot-vector determined at the start of the episode was presented (dummy).
- Input attribute 2: one-or-zero-hot-vector
  It is a one-hot-vector during the sample and target presentation periods and a zero vector during the rest of the time.
  If the start cue was [0, 1], it took the value used for matching. Otherwise, the one-hot-vector determined at the start of the episode was presented (dummy).
- Answer: a one-or-zero-hot-vector indicating whether the sample and target values matched
  [0, 0] except for the answer period, where
  [1, 0] if they matched, [0, 1] otherwise.

**Perplexity of inputs and actions**: The perplexity of possible input and target is shown below; all of the cases must be experienced to gain full knowledge.

The perplexity in the environment: Task Switch x Dummy Values x Sample Values x Target Values = 2 x 2 x 2 x 2 = 16.

The environment randomly shuffled the setting to cover all the cases within 16 episodes.

As there were two possible targets of attention in sample and target presentation periods, the perplexity of possible input and attention was 16 x 2 x 2 = 64.

## 5. Results

Experiments were conducted with the number of cells 200, 300, 400, and 500, with 10 trials of 200 episodes.

The result (number) of an episode was one of the following.

Positive: the answer was predicted correctly (Fig. 1).

Negative: the answer was not predicted correctly (Fig. 3).

Zero: no hypothesis was found or established for prediction (Fig. 2)

In many cases the task was solved in about a hundred trials with 400 cells (Fig. 1). The expected value was 0.5 as the values to be predicted were binary values. The reason why the initial result was less than 0.5 is because no hypothesis was yet formed.

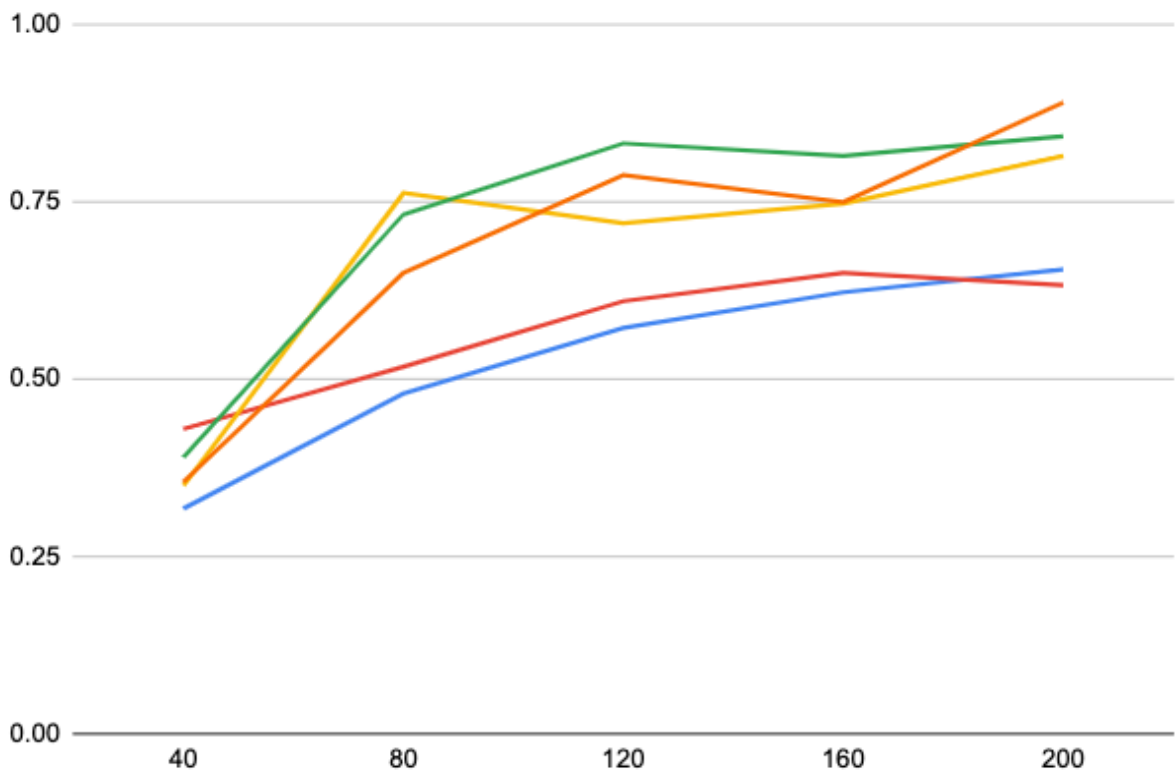

Fig. 1 Successful hypotheses (positive)

Cells : Blue 100, Red 200, Yellow 300, Green 400, Orange 500

X-axis presents episodes (average of 10 trials)

The result seems reasonable considering that the perplexity is 64 and the number of cells required to store one sequence is 4. In some cases, all the correct hypotheses could not be found even if the number of cells or the number of trials was increased further (Fig. 2). This may have been caused by a tradeoff between the randomness of attentional selection and perseverance to hypotheses.

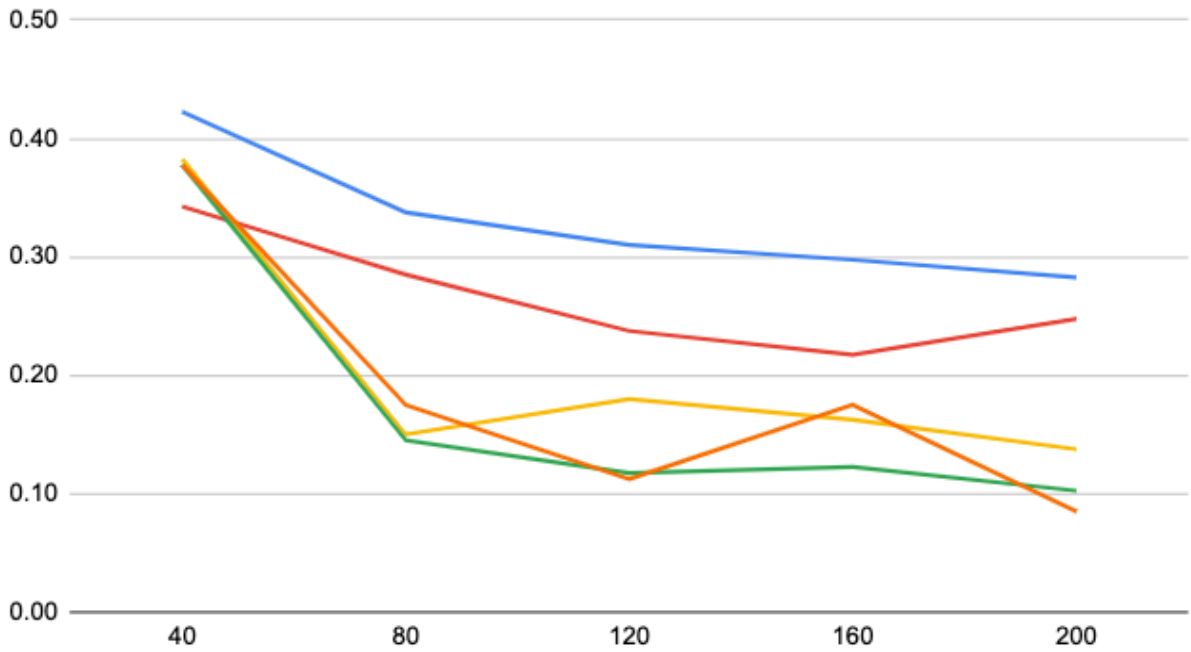

Fig. 2 No hypothesis found (Zero)

Cells : Blue 100, Red 200, Yellow 300, Green 400, Orange 500

X-axis presents episodes (average of 10 trials)

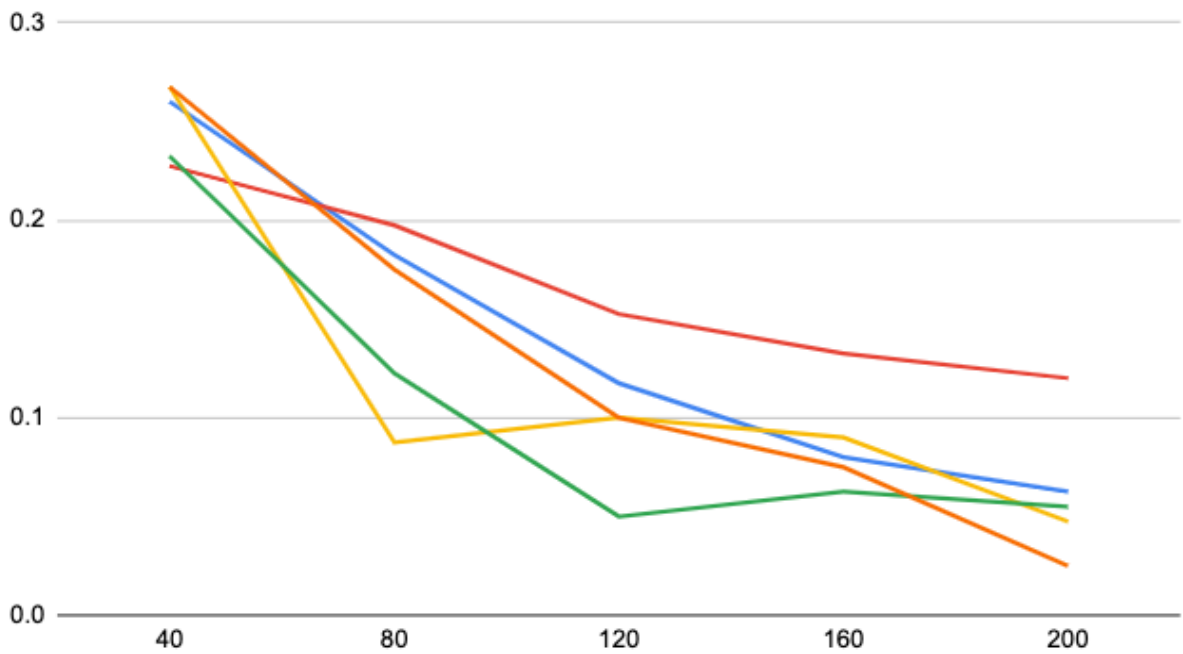

Fig. 3 Unsuccessful hypotheses (negative)

Cells : Blue 100, Red 200, Yellow 300, Green 400, Orange 500

X-axis presents episodes (average of 10 trials)

Table 1: the result in the figures

| Cell # | 100 | | | 200 | | | 300 | | | 400 | | | 500 | | |
|---|---|---|---|---|---|---|---|---|---|---|---|---|---|---|---|
| Eps. | + | - | 0 | + | - | 0 | + | - | 0 | + | - | 0 | + | - | 0 |
| 40 | 0.32 | 0.26 | 0.42 | 0.43 | 0.23 | 0.34 | 0.35 | 0.27 | 0.38 | 0.39 | 0.23 | 0.38 | 0.36 | 0.27 | 0.38 |
| 80 | 0.48 | 0.18 | 0.34 | 0.52 | 0.20 | 0.29 | 0.76 | 0.09 | 0.15 | 0.73 | 0.12 | 0.15 | 0.65 | 0.16 | 0.18 |
| 120 | 0.57 | 0.12 | 0.31 | 0.61 | 0.15 | 0.24 | 0.72 | 0.1 | 0.18 | 0.83 | 0.05 | 0.12 | 0.79 | 0.1 | 0.11 |
| 160 | 0.62 | 0.08 | 0.30 | 0.65 | 0.13 | 0.22 | 0.75 | 0.09 | 0.16 | 0.82 | 0.06 | 0.12 | 0.75 | 0.06 | 0.18 |
| 200 | 0.66 | 0.06 | 0.28 | 0.63 | 0.12 | 0.25 | 0.82 | 0.05 | 0.14 | 0.84 | 0.06 | 0.10 | 0.89 | 0.03 | 0.09 |

An example of an episode is shown in the Appendix.

The implementation of the agent and environment can be found on GitHub:

https://github.com/rondelion/SMA4DM2S_NN

## 6. Discussion

The proposed implementation is yet to be improved in terms of problem-solving performance. For one thing, while match-to-sample tasks for human beings normally use figures as input, low-dimensional vectors were used as input in this report. Visual analogy tasks often used for testing fluid intelligence also use figures as input. In order to deal with graphic input, an image recognition mechanism is required. For another thing, the present proposal is based on (rote) memorization of sequences using neural circuits and does not generalize patterns. Pattern generalization is usually performed by pattern recognition learning. As a point related to generalization, for the proposed mechanism to attend to only one attribute at a time, it cannot deal with conjunction of attributes.

Now, let us note the relationship between the proposed model and reinforcement learning. The proposed model does not assume a Markov decision process (MDP) in which the input is observed as states, for the input sequence is deterministic and different input sequences may follow the same input sequence, as noted in the sequence memory section. However, as the transitions of the internal states were described by a transition matrix, it can be described as a deterministic partially observed Markov decision process (POMDP). While solving a POMDP normally requires a considerable mathematical setup (Kaelbling, 1998), the present model solves the task for a relatively short deterministic sequence with a simple mechanism, by making the internal state deterministically mapped from the input sequence. While the evaluation of the internal state in ordinary reinforcement learning is propagated within internal states and requires a large number of trials, the present model used inverse replay to evaluate the sequence, to result in reducing the number of trials.

Planning could be done using a mechanism similar to the one proposed here. In planning, trials are not conducted by interaction with the environment, but by simulation within the agent. An agent with a sequence memory mechanism would remember failed trials in a simulation, and (probabilistically) avoid action within failed trials. If a successful sequence is found in the simulation, it is replayed and executed.

Sequence memory can help solve the binding problem of distributed representations (O'Reilly, 2002). In a distributed representation such as a neural network, the place to represent an attribute such as the color of a figure tends to be fixed, and the problem is how to represent and relate the attributes of, e.g., more than one figure. (In a symbolic representation system such as a computer language, the problem does not arise because different objects can be represented in different places.) In sequence memory, objects experienced at different times are associated with different internal states, so that objects can be distinguished by differences in internal states (Arakawa, 2015). Objects located in different places can also be associated with different internal states by serializing them through a mechanism such as eye movement. Note that to represent information on what an object is and where it is located separately could also help solve the binding problem as in the brain.

Since the fluid intelligence task involves conscious and sequential processing, System 2 in dual-process theory (Kahneman, 2003) is thought to be involved. However, the overall processes in the current model were parallel and automatic; in particular, the recall of zero or more past sequences was parallel. Action selection was also automatic from parallel sequence recall. Thus, the model would not be a model of System 2, though "conscious" (System 2-like) replay of sequences, which was not dealt in this model, may contribute to efficient information processing.

Now, let us discuss the biological plausibility of the model briefly, though it was stated in the Introduction that the proposed model does not aim at precise biological plausibility. In order for the model to be biologically plausible, it must map to neuroscientific facts. First, let us look at the correspondence between the five assumptions made in the Introduction. For Assumption 1) "the agent has an input-output sequence and a corresponding internal state sequence," it is known that replay/preplay is performed at least in the hippocampus (Dragoi, 2011). Assumption 2) "the agent can recall past internal state and input-output sequences from the current input sequence" also corresponds to hippocampal sequence memory. For Assumption 3) "the agent can replay the internal state sequence in reverse when a reward is given, in order to evaluate the sequence," reverse replay in the hippocampus has been reported (Foster, 2006), though the mechanism for reinforcement is not known. For Assumption 4) "the agent probabilistically determines the output using parallelly recalled internal state sequences and the values associated to the sequences" assumes Assumption 3, and it further raises the issue of whether internal states are activated in parallel (incidentally, the neural network operates essentially in parallel). In the hippocampus, where associations of sequences are thought to take place, complex processing including EEG synchronization is taking place, and if we were to map it to the mechanism proposed here, it would remain at an abstract level. In addition, given the findings that the prefrontal cortex is responsible for rule discovery (Badre, 2010)(Cao, 2006)(Miller, 2001) and that it is the prefrontal cortex that actually performs output (action), the coordination between the hippocampus and prefrontal cortex should be considered. For Assumption 5) "the agent pays attention to a specific attribute in the input at a time and remembers only the selected attribute" is based on the selective attention hypothesis (van Moorselaar, 2020), while there are various theories about its neural mechanism. As described in the Sequence Memory section, the implementation method of sequence memory is not an essential part of the proposal, and its biological plausibility should be questioned only with respect to the above assumptions.

Finally, let us note the similarities and differences between the present model and transformer (Polosukhin, 2017). The similarities are that both deal with sequences, "pay attention" to the input attributes at time points, predict the sequence, and learn behavior through prediction. The differences are that a) in transformer, attention is continuous, whereas in this model it is binary, b) in transformer, attention is (basically) determined from the entire episode, whereas in this model attention is determined from the sequence up to a time point, and c) transformer is a differential learning system, whereas this model uses a simpler reinforcement method. While it may be possible to create a model that shows behavior similar to the present model by applying the transformer algorithm, it will be left for another study.

# 7. Conclusion

This report proposed a cognitive model for discovering regularity of event sequences based on sequence memory. The reason for modeling the function of finding the regularity of event sequences based on sequence memory is that it would be effective in fluid intelligence tasks such as visual analogy and match-to-sample tasks to find regularities by memorizing multiple event sequences from the task. Since fluid intelligence tasks require a relatively small number of event sequences to find a solution, a memory model instead of a statistical learning model was used. In modeling, a relatively simple neural network model was created with abstract biological plausibility in mind. To test the model, a delayed match-to-sample task was used. The proposed model showed reasonable performance for the complexity of the task.

# Appendix: Learning in Action with a Simple Delayed Match-to-Sample Task

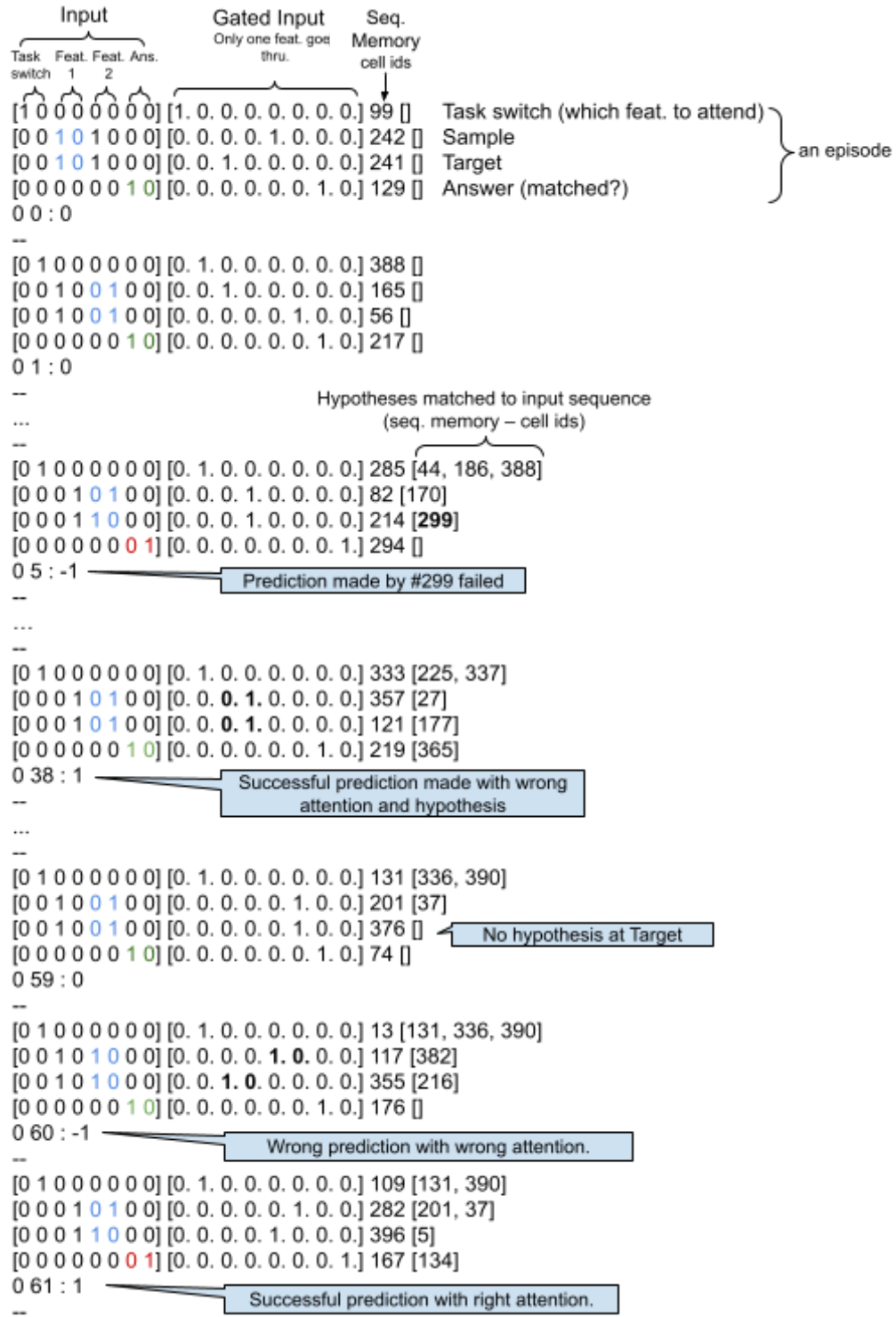

**Fig. 4**